\documentclass[letterpaper, 10 pt, conference]{ieeeconf}  
\usepackage{amsmath,amsfonts}
\usepackage{algorithmic}
\usepackage{algorithm}
\usepackage{booktabs}
\usepackage{multirow}
\usepackage{array}
\usepackage{tabularx}
\usepackage{subcaption}
\usepackage{textcomp}
\usepackage{gensymb}
\usepackage{stfloats}
\usepackage{verbatim}
\usepackage{graphicx}
\PassOptionsToPackage{hyphens}{url}
\usepackage{xurl}
\usepackage{pifont}
\usepackage[colorlinks=true, citecolor=blue, linkcolor=blue, urlcolor=blue, breaklinks=true]{hyperref}

\IEEEoverridecommandlockouts               

\title{\LARGE \bf
HydroMap: Probabilistic Water Surface Elevation Mapping for
Semantic Scene Representation in Inland Waterways
}

\author{Zhongbi Luo$^{1}$, Yunjia Wang$^{2,3}$, Herman Bruyninckx$^{1,3,4}$, and Peter Slaets$^{1}$%
\thanks{Corresponding author: Zhongbi Luo (e-mail: \href{mailto:zhongbi.luo@kuleuven.be}{zhongbi.luo@kuleuven.be}).}%
\thanks{$^{1}$Zhongbi Luo, Herman Bruyninckx, and Peter Slaets are with the Division of Robotics, Automation and Mechatronics, Department of Mechanical Engineering, KU Leuven, 3001 Leuven, Belgium (e-mail: zhongbi.luo@kuleuven.be; herman.bruyninckx@kuleuven.be; peter.slaets@kuleuven.be).}%
\thanks{$^{2}$Yunjia Wang is with the Division of Declarative Languages and Artificial Intelligence (DTAI), Department of Computer Science, KU Leuven, 8200 Bruges, Belgium (e-mail: yunjia.wang@kuleuven.be).}%
\thanks{$^{3}$Yunjia Wang and Herman Bruyninckx are also with Flanders Make@KU Leuven, 3001 Leuven, Belgium.}%
\thanks{$^{4}$Herman Bruyninckx is also with the Department of Mechanical Engineering, TU Eindhoven, 5612 AZ Eindhoven, The Netherlands.}%
}

\begin{document}

\maketitle
\thispagestyle{empty}
\pagestyle{empty}

\begin{abstract}

Autonomous surface vehicles operating in inland waterways can benefit from a persistent representation of both surrounding structures and the water surface. LiDAR-based simultaneous localization and mapping often produces sparse or missing water returns, leaving this operational surface absent from the reconstructed scene. We propose HydroMap, an odometry-decoupled framework that reconstructs water surface elevation from stereo observations and integrates it with the structural map. Per-frame water points form joint cell observations with propagated stereo and pose uncertainty, and successive observations are fused into a persistent probabilistic elevation map. Semantic map conversion then combines the elevation map with structural geometry in a unified 2.5D representation of water, boundaries, structures, and overhead regions. Across three verified LiDAR reference regions in the Pohang Canal and Leuven Vaart datasets, the elevation root mean square error remains below  $5$\,cm relative to references expressed in the same map frame. The elevation and semantic maps are published at $2$\,Hz and $1$\,Hz, respectively. HydroMap thereby complements LiDAR maps with a persistent representation of the water surface for downstream navigation in inland waterways.

\end{abstract}

\section{INTRODUCTION}


Autonomous surface vehicles (ASVs) are increasingly deployed in inland waterways, where the navigable space is constrained by surrounding structures~\cite{wang2023roboat}. Typical operational scenarios include passing beneath bridges with limited vertical clearance, transiting locks with varying water levels, navigating narrow channels bounded by quay walls and vegetation, and berthing alongside fixed structures~\cite{kim2023navigable,luo2025inland}. Safe operation in these settings depends not only on perceiving the surrounding structures, but also on knowing \textit{where the water surface lies}. The water surface determines the clearance beneath overhead obstacles and establishes the spatial reference between the vessel and the built environment. This requires a three-dimensional map that captures both the surrounding infrastructure and the water surface itself.

\begin{figure}[!t]
  \centering
  \includegraphics[width=0.99\columnwidth]{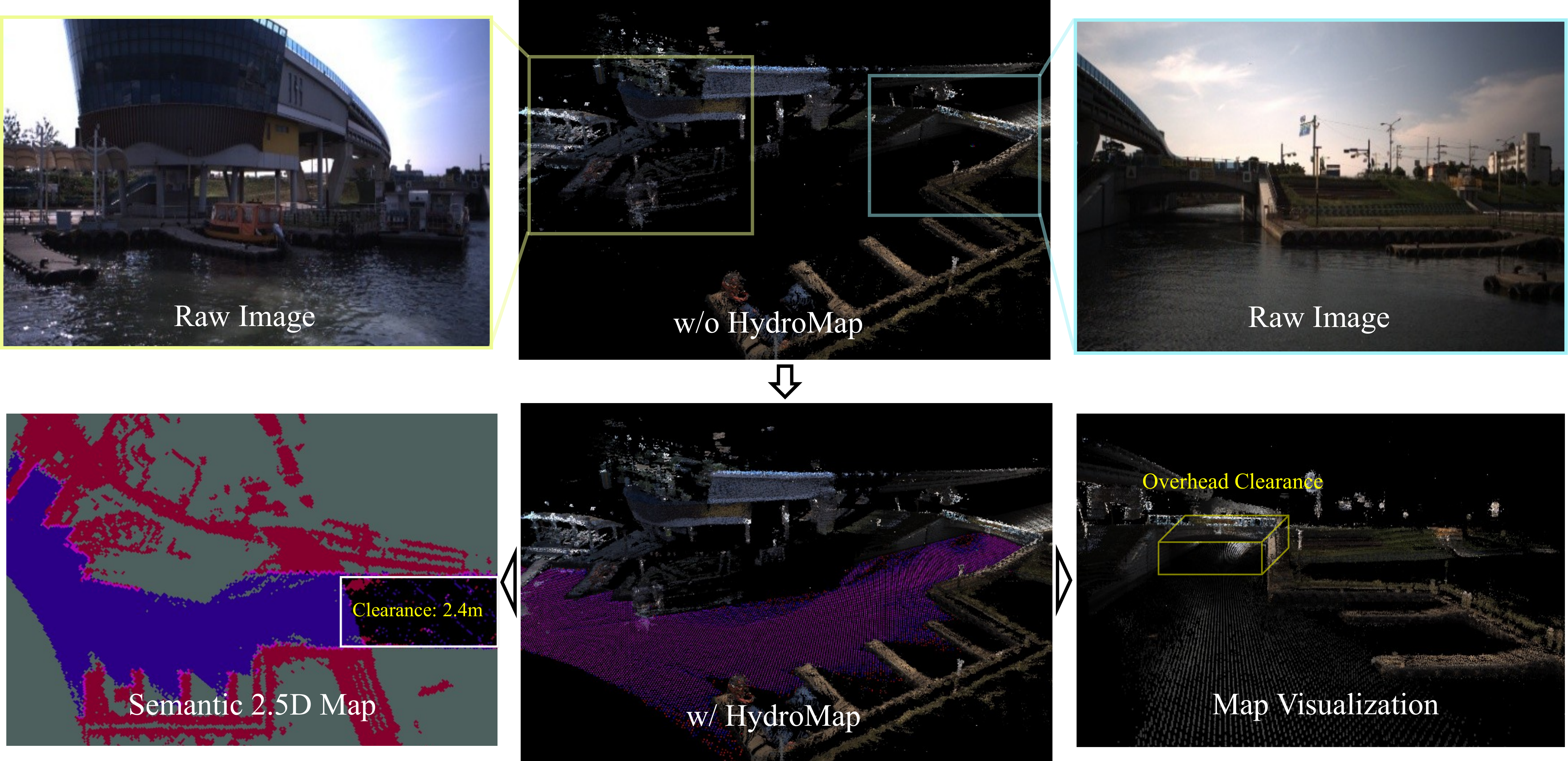}
  \caption{Qualitative comparison on the Pohang Canal dataset~\cite{chung2023pohang}. The top row shows representative images from the left camera and the map reconstructed by FAST-LIVO2~\cite{zheng2024fast} without HydroMap. The bottom row shows the map reconstructed with HydroMap, together with a 3D map visualization with labeled overhead structures and the corresponding semantic 2.5D map after map conversion.}
  \label{fig:first}
\end{figure}

Simultaneous localization and mapping (SLAM) has made significant progress toward this goal, and recent LiDAR-based systems such as FAST-LIO2~\cite{xu2022fast} achieve centimeter-level accuracy in structured environments. The number of valid LiDAR returns from water, however, depends on the incidence angle, the sensor-to-surface distance, and the water turbidity~\cite{ranieri2024water,allis2011application}; under typical inland conditions most pulses are absorbed and few or no returns are produced. LiDAR point clouds from~\cite{chung2023pohang,luo2026leuvenvaartdataset} also reveal mirror reflections near the water-structure boundary, which place symmetric copies of nearby structures below the water plane. The vessel's operational surface therefore appears as a void in the reconstructed map---the ASV can neither delineate the water surface boundary, nor assess whether sufficient clearance exists beneath overhead obstacles such as bridges.

To address this gap, prior work has explored water surface perception through different sensing modalities. Semantic segmentation of images has been used to detect the navigable area~\cite{hammedi2022reliable, yao2021shorelinenet}, while stereo vision has been employed to reconstruct the 3D water surface from individual frames~\cite{bergamasco2017wass}. Visual water perception is affected by low texture, reflections, and sun glint. The former provides 2D navigable boundaries without surface elevation, and the latter produces only local per-frame estimates that may contain holes or erroneous matches. Neither constructs a persistent water surface map in a common scene frame.

In this paper, we present \textbf{HydroMap}, as shown in Fig.~\ref{fig:first}. HydroMap is a mapping framework designed to fill the water surface void left by LiDAR-based SLAM systems in inland waterways. The key idea is to reconstruct the water surface as a probabilistic elevation map from stereo vision and integrate it with the existing non-water point cloud map to form a more complete 3D scene representation, from which water surface boundaries and overhead clearance can be directly derived.  Notably, the framework is \textit{odometry-decoupled}: it requires only gravity estimates, poses, and associated covariances from an external odometry source.

The main contributions of this work are summarized as follows: 
\begin{itemize} 
\item An odometry-decoupled mapping system that integrates stereo water surface perception, persistent probabilistic elevation mapping, and semantic integration with the structural LiDAR map.

\item A probabilistic elevation mapping method that forms joint cell observations, propagates stereo and pose uncertainty, and uses Split Covariance Intersection (SCI)~\cite{julier2001simultaneous,li2013split} to fuse successive observations without assuming independent pose-induced errors across frames.

\item A semantic scene representation that combines the water surface with surrounding structures in a semantic OctoMap~\cite{hornung13auro} and a 2.5D grid, with field validation on Leuven Vaart and the public Pohang Canal dataset~\cite{chung2023pohang}.
\end{itemize}

\begin{figure*}[!t]
  \centering
  \vspace*{8pt}
  \includegraphics[width=0.70\textwidth]{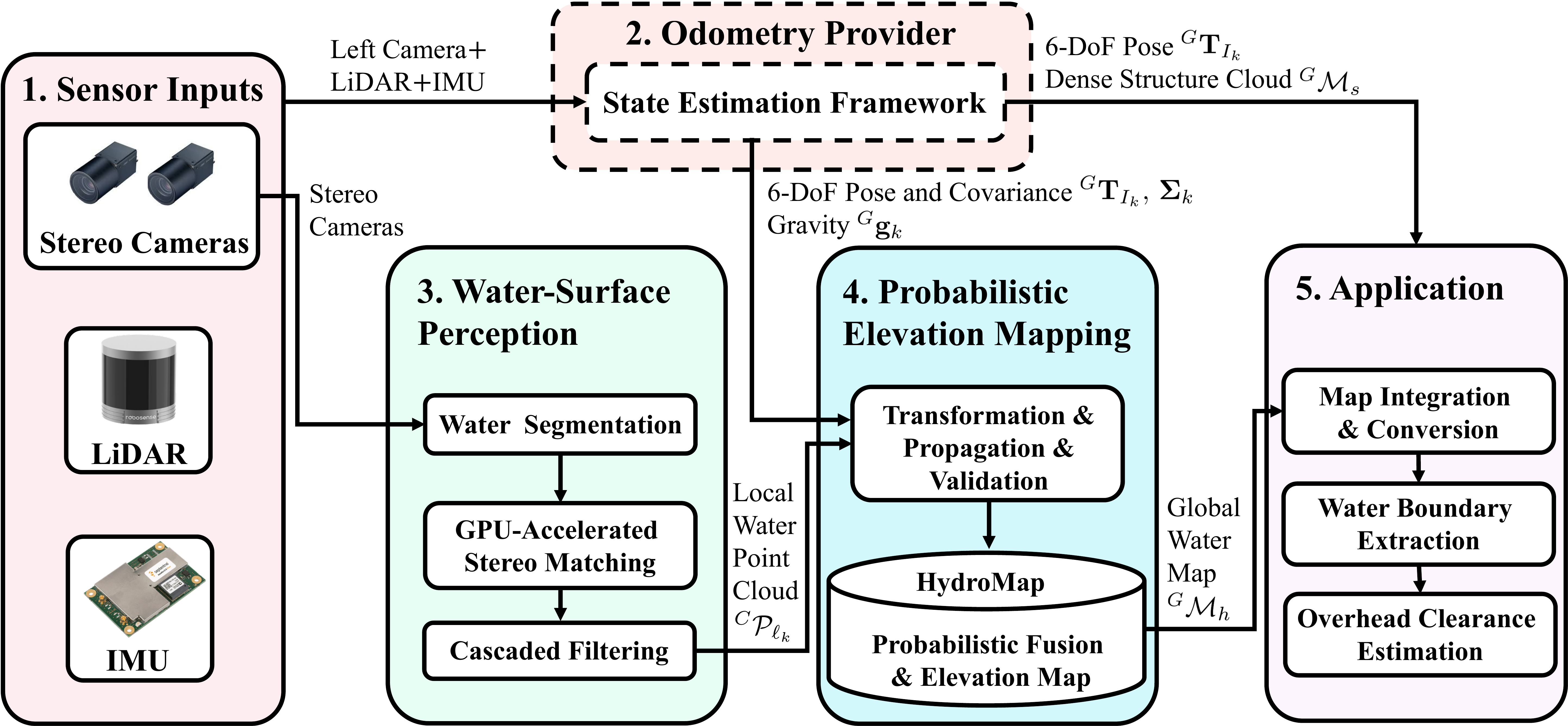}
  \caption{Overview of the proposed HydroMap system.}
  \label{fig:system_architecture}
  \vspace*{-8pt}
\end{figure*}
\section{RELATED WORK}
\subsection{Visual Water Surface Perception for ASV Navigation}

Because LiDAR sensing degrades severely over water surfaces~\cite{ranieri2024water,allis2011application}, vision-based methods have emerged as an important approach to water surface perception on ASVs.

Semantic segmentation has been widely adopted to detect navigable water regions in camera images~\cite{hammedi2022reliable,yao2021shorelinenet}. Kim et al.~\cite{kim2023navigable} projected the segmented region onto an ASV-fixed water plane and extracted boundaries for local planning. This 2D representation does not provide the vertical geometry required for overhead clearance estimation.

Stereo vision provides metric depth and has been applied to water surface reconstruction~\cite{bergamasco2017wass,benetazzo2016stereo}. Griesser et al.~\cite{griesser2023visual} restricted semi-global matching (SGM)~\cite{hirschmuller2008stereo} to water pixels, reconstructed a water point cloud, and fitted a plane to estimate vessel attitude. These results establish that stereo can recover water geometry, but per-frame estimates remain local and may be discontinuous under textureless or reflective conditions. A natural way to mitigate these effects is to accumulate repeated observations over time. As the ASV moves, the same water region is observed multiple times from slightly different viewpoints. Consequently, observation gaps in a single frame can be filled with valid measurements from subsequent frames.

\subsection{Probabilistic Elevation Mapping and Semantic Scene Representation}
Representing terrain with quantified uncertainty has been well studied in ground robotics~\cite{belter2012estimating, fankhauser2018probabilistic}. Fankhauser et al.~\cite{fankhauser2018probabilistic} proposed a probabilistic elevation mapping framework in which per-cell height estimates are updated with a Kalman filter, and robot pose uncertainty is propagated to obtain terrain estimates with associated confidence bounds.

GPU implementations~\cite{miki2022elevation} further demonstrate real-time probabilistic elevation mapping. Conventional Covariance Intersection (CI) provides a consistent fusion rule when cross-correlation is unknown~\cite{julier1997non}. In HydroMap, stereo measurement uncertainty is propagated for each observation, while uncertainty from the shared odometry pose is handled separately because pose errors may be correlated across successive frames. SCI preserves this separation and enables temporal fusion without requiring the unavailable cross-covariance between successive pose errors~\cite{julier2001simultaneous,li2013split}.

Dense 3D maps are also projected into per-cell height and slope representations for efficient navigation queries~\cite{fredriksson2024voxel}. Related projections have been used in inland waterways for shoreline and bridge detection~\cite{luo2025inland}.

These approaches, however, require the traversable surface to be sufficiently observed by range sensors, a condition often not met in inland waterways. This motivates modeling the water surface as an uncertain elevation map inferred from stereo observations and integrating it with the structural geometry in a persistent semantic scene representation.

\section{SYSTEM OVERVIEW}

\subsection{Notation and Coordinate Frames}
\label{sec:notation}

In this work, we define four coordinate frames: the global frame $G$, the IMU/body frame $I$, the left camera frame $C$, and the LiDAR frame $L$. The global frame $G$ is initialized at the first body frame. For instance, the extrinsic transformation from the camera frame to the body frame is denoted by ${}^{I}\mathbf{T}_{C} \in SE(3)$. All sensors are rigidly attached; their extrinsics are obtained via offline calibration, and their measurements are time-synchronized.

At timestamp $t_k$, the odometry state is represented by the pose of the body in the global frame ${}^{G}\mathbf{T}_{I_k}=[\,{}^{G}\mathbf{R}_{I_k}\mid{}^{G}\mathbf{p}_{I_k}\,]$, where ${}^{G}\mathbf{R}_{I_k}\in SO(3)$ is the rotation and ${}^{G}\mathbf{p}_{I_k}\in\mathbb{R}^3$ is the position. The state also includes the estimated gravity vector ${}^{G}\mathbf{g}_k\in\mathbb{R}^3$ and the pose covariance $\boldsymbol{\Sigma}_k$.

\subsection{System Pipeline}
\label{sec:pipeline}

Fig.~\ref{fig:system_architecture} illustrates the HydroMap pipeline. At each stereo timestamp $t_k$, HydroMap takes synchronized stereo images together with the externally provided ${}^{G}\mathbf{T}_{I_k}$, ${}^{G}\mathbf{g}_k$, and $\boldsymbol{\Sigma}_k$. The water surface perception module (Sec.~\ref{sec:perception}) reconstructs a local water point cloud ${}^{C}\mathcal{P}_{\ell_k}$; the elevation mapping module (Sec.~\ref{sec:elevation}) transforms it into the global frame, validates it against ${}^{G}\mathbf{g}_k$, and fuses it into the persistent probabilistic water surface elevation map ${}^{G}\mathcal{M}_h$; and the application stage (Sec.~\ref{sec:application}) integrates ${}^{G}\mathcal{M}_h$ with the structural map ${}^{G}\mathcal{M}_s$ to extract the water boundary and estimate the overhead clearance.

\section{METHODOLOGY}
\label{sec:method}

\subsection{Odometry Provider}
\label{sec:odom}

HydroMap uses the output of an external odometry system. The probabilistic elevation mapping module requires gravity, pose, and pose covariance at each $t_k$. This module is not used to update the odometry state.
In this work, these quantities are provided by FAST-LIVO2~\cite{zheng2024fast}, which performs tightly coupled LiDAR-inertial-visual state estimation and reports a state covariance matrix together with the pose. It makes effective use of the sensing modalities already available on the platform, and its visual constraints complement LiDAR and IMU measurements in challenging inland environments. The structural map maintained by FAST-LIVO2 is further used as $\mathcal{M}_s$ in the downstream application stage. The provider and HydroMap run as separate processes that exchange only timestamped messages, so the mapping workload stays outside the state estimation loop.

\subsection{Water Surface Perception}
\label{sec:perception}
\begin{figure}[!t]
  \centering
  \vspace*{8pt}
  \includegraphics[width=0.82\columnwidth]{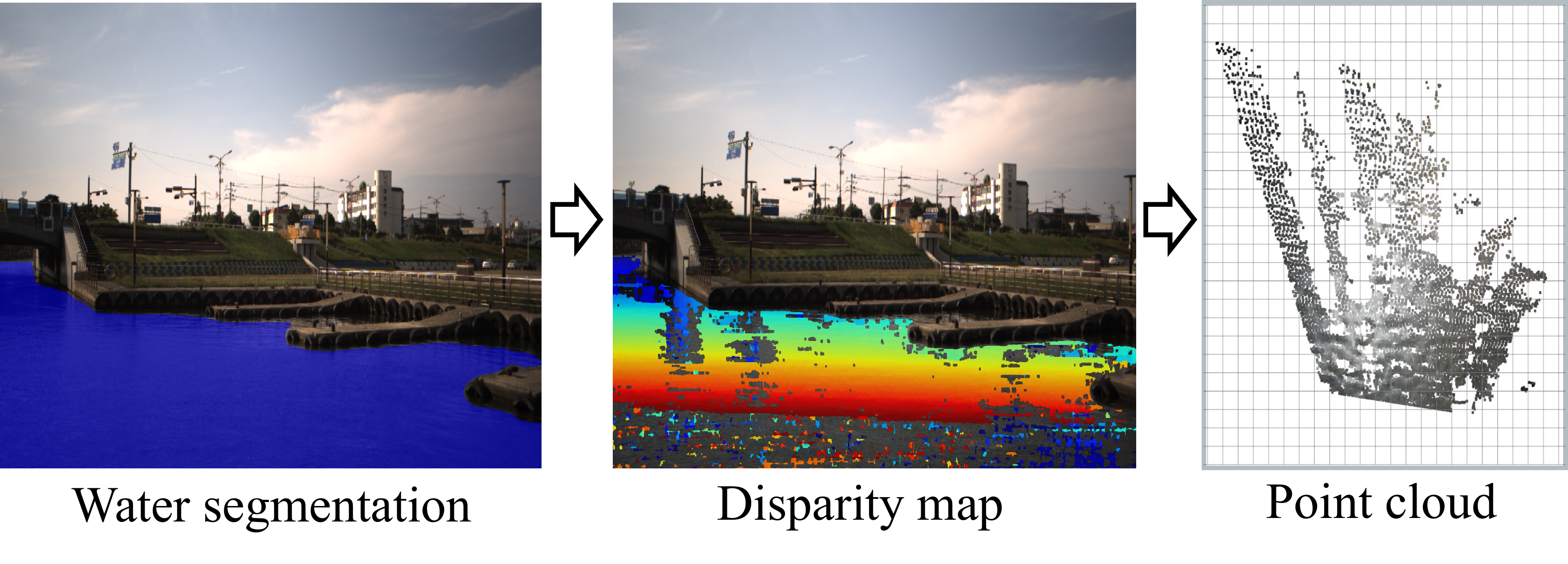}
  \caption{Overview of the water surface perception pipeline. From left to right: raw image overlaid with the water segmentation mask, the disparity map in water regions, and the local water point cloud visualized from a bird's-eye view.}
  \label{fig:stereo_pipeline}
  \vspace*{-8pt}
\end{figure}

This module extracts a local water surface point cloud ${}^{C}\mathcal{P}_{\ell_k}$ from a calibrated stereo pair at each timestamp $t_k$. The pipeline consists of three stages: water segmentation, GPU-accelerated stereo matching, and cascaded geometric filtering. The result of each stage is shown in  Fig.~\ref{fig:stereo_pipeline}.

\subsubsection{Water Segmentation}

Water surface segmentation is challenging because low texture and strong reflections of the sky, shoreline vegetation, and surrounding structures create substantial appearance variation and ambiguous water--shore boundaries. We therefore fine-tune a DeepLabV3+ network~\cite{chen2018encoder} with a ResNet-50 backbone~\cite{he2016deep}, initialized with PASCAL VOC-pretrained weights, using the manually annotated inland waterway images described in Sec.~\ref{sec:seg_performance}. At timestamp $t_k$, the network processes the rectified left image $\mathbf{I}_{L_k}$ and outputs a binary mask $\mathcal{W}_k$, where $\mathcal{W}_k(u,v)=1$ denotes water and $\mathcal{W}_k(u,v)=0$ denotes background. The mask restricts stereo matching to water pixels and suppresses false matches elsewhere.

\subsubsection{GPU-Accelerated Stereo Matching}
 
The rectified stereo pair is processed with the eight-path GPU implementation of SGM~\cite{hirschmuller2008stereo} in libSGM~\cite{libSGM}. The estimated disparity $d_k$ is masked by $\mathcal{W}_k$:
\begin{equation}
  d_{w_k}(u,v) =
  \begin{cases}
    d_k(u,v) & d_k(u,v) > d_{\min},\ \mathcal{W}_k(u,v) = 1, \\
    \text{invalid} & \text{otherwise,}
  \end{cases}
  \label{eq:masked_disp}
\end{equation}
where $d_{\min}=0$ is the minimum valid disparity. Morphological filtering closes small gaps in $d_{w_k}$.

\subsubsection{Cascaded Geometric Filtering}

The cleaned disparity is reprojected into a depth map $Z_k$ in the left camera frame $C$. A cascaded filter keeps depths between 0 and 50 m, removes gradients above the $98$th percentile~\cite{bergamasco2017wass}, and selects the three largest 4-connected components. RANSAC plane fitting~\cite{fischler1981random} then selects points within 0.1 m of the dominant surface. The resulting local cloud ${}^{C}\mathcal{P}_{\ell_k}=\{{}^{C}\mathbf{p}^{(n)}_{k}\}_{n=1}^{N_k}$ is passed to probabilistic elevation mapping.

\subsection{Probabilistic Elevation Mapping}
\label{sec:elevation}
The probabilistic elevation mapping module fuses successive, local water surface observations into a persistent global map ${}^{G}\mathcal{M}_h$. Because the water surface perception and odometry modules run asynchronously, HydroMap synchronizes them by maintaining a first-in-first-out buffer of incoming water surface point clouds and a history of odometry states. For a given observation at $t_k$, the corresponding pose ${}^{G}\mathbf{T}_{I_k}$ and its covariance $\boldsymbol{\Sigma}_k$ are obtained by interpolating the odometry states, using spherical linear interpolation (SLERP)~\cite{shoemake1985animating} for orientation and linear interpolation for position and covariance.

To incrementally fuse these multi-frame observations, the map ${}^{G}\mathcal{M}_h$ is structured as a hash-indexed 2D grid with a resolution of $\Delta_g = 0.3$\,m. Each cell $i$ maintains an elevation estimate and a split variance with an independent component $p_{u,i}$ and a potentially correlated component $p_{c,i}$:
\begin{equation}
  h_i \sim \mathcal{N}(\hat{h}_i,p_{u,i}+p_{c,i}).
  \label{eq:cell_belief}
\end{equation}
The fusion process consists of three main steps: per-point global transformation and uncertainty propagation, per-frame validation, and joint per-cell fusion.

\subsubsection{Per-Point Global Transformation and Uncertainty Propagation}
The local point cloud ${}^{C}\mathcal{P}_{\ell_k}$ is transformed into the body (IMU) frame $I$ to align with the external odometry state. For a point ${}^{C}\mathbf{p}^{(n)}_k$, its coordinate in the body frame is obtained via the sensor extrinsics: ${}^{I}\mathbf{p}^{(n)}_k = {}^{I}\mathbf{R}_C {}^{C}\mathbf{p}^{(n)}_k + {}^{I}\mathbf{p}_C$. Its global coordinate is subsequently computed via the estimated body pose: ${}^{G}\mathbf{p}^{(n)}_k =
{}^{G}\mathbf{R}_{I_k}\,
{}^{I}\mathbf{p}^{(n)}_k + {}^{G}\mathbf{p}_{I_k}.$

Stereo reconstruction error is dominated by depth uncertainty, which increases rapidly with distance. For a rectified stereo pair with focal length $f_x$ and baseline $B$, the depth satisfies $Z=f_x B/d$. A first-order propagation of the disparity noise with standard deviation $\sigma_d$ (in pixels) yields a depth variance that grows with the fourth power of depth:
\begin{equation}
\sigma_{Z,n}^{2}
=
\left(\frac{\partial Z}{\partial d}\right)^{\!2} \sigma_{d}^{2}
=
\frac{\sigma_{d}^{2}}{f_x^{2} B^{2}}\,Z_{n}^{4},
\end{equation}
where $Z_n = {}^{C}p^{(n)}_{k,z}$ is the depth of the point from the camera. We construct the simplified sensor covariance in the camera frame using a diagonal form:
\begin{equation}
{}^{C}\boldsymbol{\Sigma}^{(n)}_{\ell}
\approx
\mathrm{diag}\!\left(
  \sigma_{xy}^{2},\,
  \sigma_{xy}^{2},\,
  \sigma_{Z,n}^{2}
\right),
\end{equation}
where $\sigma_{xy}^{2}$ is a small constant lateral variance that accounts for residual calibration effects and lateral pixel uncertainty. This covariance is then rotated into the body frame using:
\begin{equation}
{}^{I}\boldsymbol{\Sigma}^{(n)}_{\ell} =
{}^{I}\mathbf{R}_{C} \,
{}^{C}\boldsymbol{\Sigma}^{(n)}_{\ell} \,
{}^{I}\mathbf{R}_{C}^{\top}.
\end{equation}

The elevation error is separated into the stereo uncertainty associated with each point and the pose uncertainty shared by all points observed at $t_k$. The stereo term is the vertical component of the rotated covariance,
\begin{equation}
\sigma_{z,n}^{2} = \mathbf{e}_z^{\top}\,{}^{G}\mathbf{R}_{I_k}\,{}^{I}\boldsymbol{\Sigma}^{(n)}_{\ell}\,{}^{G}\mathbf{R}_{I_k}^{\top}\mathbf{e}_z, \quad \mathbf{e}_z = [0,0,1]^{\top}.
\end{equation}

The pose term is obtained from the sensitivity of the elevation to $\boldsymbol{\delta}_k = [\,\delta\mathbf{p}^{\top},\,\delta\boldsymbol{\theta}^{\top}]^{\top}$, with covariance $\boldsymbol{\Sigma}_k$. The position error is expressed in the global frame and the rotation error as a small angle perturbation in the body tangent space. The pose error shifts the elevation of point $n$ by $\mathbf{J}^{(n)\top}_{k}\boldsymbol{\delta}_k$, with
\begin{equation}
\mathbf{J}^{(n)}_{k} = \bigl[\, \mathbf{e}_z^{\top},\; -\mathbf{e}_z^{\top}\,{}^{G}\mathbf{R}_{I_k}\lfloor {}^{I}\mathbf{p}^{(n)}_{k} \rfloor_{\wedge} \,\bigr]^{\top} \in \mathbb{R}^{6},
\end{equation}
where $\lfloor \cdot \rfloor_{\wedge}$ denotes the skew-symmetric matrix associated with the cross product, and ${}^{I}\mathbf{p}^{(n)}_k$ is the lever arm from the body origin to the point. The full $6\times6$ marginal pose covariance $\boldsymbol{\Sigma}_k$, including the cross-covariance between translation and rotation, is available at each timestamp.

\subsubsection{Per-Frame Validation}
Before fusion, each frame undergoes three validation checks operating on the water surface point cloud $\{{}^{G}\mathbf{p}^{(n)}_k\}_{n=1}^{N_k}$. \textit{Support gating} rejects frames with fewer than $N_{\min} = 300$ points, which carry too little evidence for the aggregation described below. \textit{Area gating} rejects frames whose horizontal bounding-box area is smaller than $A_{\min}$ (set to $20\,\mathrm{m}^2$ in our implementation), a condition that typically indicates a partial segmentation failure, such as detecting only a thin sliver of water along a reflective hull or shoreline. \textit{Tilt gating} rejects frames whose reconstructed surface normal is inconsistent with the gravity vector. Let $\hat{\mathbf{n}}_k$ be the surface normal estimated via principal component analysis (PCA) on the global water surface point cloud. A frame is accepted only if
\begin{equation}
  \alpha_k = \arccos\!\bigl( |\hat{\mathbf{n}}_k \cdot \hat{\mathbf{e}}_{g,k}| \bigr) < \alpha_{\max}, \quad \hat{\mathbf{e}}_{g,k} = \frac{{}^{G}\mathbf{g}_k}{\|{}^{G}\mathbf{g}_k\|},
\end{equation}
with $\alpha_{\max}$ set to $15\degree$ and ${}^{G}\mathbf{g}_k$ the gravity vector provided by the odometry state. A frame that fails any gate is dropped entirely.

\subsubsection{Joint Per-Cell Fusion}

Each validated point
$({}^{G}\mathbf{p}^{(n)}_k,\sigma_{z,n}^2,\mathbf{J}^{(n)}_k)$
is assigned to a grid cell according to its horizontal coordinates.
Its elevation measurement is
\begin{equation}
  z^{(n)}_k = \mathbf{e}_z^\top {}^{G}\mathbf{p}^{(n)}_k.
\end{equation}
Because all points observed at $t_k$ share the pose error $\boldsymbol{\delta}_k$, the $N_{ik}$ points assigned to cell $i$ form one joint observation. Their elevations, stacked into $\mathbf{z}_{ik}\in\mathbb{R}^{N_{ik}}$, follow
\begin{equation}
  \begin{aligned}
    \mathbf{z}_{ik}
      &= \mathbf{1}h_i
       + \mathbf{J}_{ik}\boldsymbol{\delta}_k
       + \mathbf{1}(m_{u,ik}+m_{c,ik})
       + \boldsymbol{\varepsilon}_{ik}, \\
    m_{u,ik} &\sim \mathcal{N}(0,\sigma_{m,u}^2), \qquad
    m_{c,ik} \sim \mathcal{N}(0,\sigma_{m,c}^2), \\
    \boldsymbol{\varepsilon}_{ik}
      &\sim \mathcal{N}(\mathbf{0},\mathbf{D}_{ik}).
  \end{aligned}
  \label{eq:joint_observation}
\end{equation}
Here, $\mathbf{1}$ is the all-ones vector,
$\mathbf{J}_{ik}\in\mathbb{R}^{N_{ik}\times6}$ stacks
$\mathbf{J}^{(n)\top}_k$ as rows, and
$\mathbf{D}_{ik}=\operatorname{diag}(\sigma_{z,n}^2)$ contains
the propagated stereo variances. The residual $m_{u,ik}$ represents
variations that are assumed independent between observations, including
small surface fluctuations and random reconstruction errors. The residual
$m_{c,ik}$ represents effects that may persist, such as camera model and
calibration errors. Within the current observation,
$\boldsymbol{\delta}_k$, $m_{u,ik}$, $m_{c,ik}$, and
$\boldsymbol{\varepsilon}_{ik}$ are modeled as mutually independent.
Marginalizing these terms with
$\boldsymbol{\delta}_k\sim
\mathcal{N}(\mathbf{0},\boldsymbol{\Sigma}_k)$ gives
\begin{equation}
  \mathbf{R}_{ik}
    = \mathbf{D}_{ik}
    + \mathbf{J}_{ik}\boldsymbol{\Sigma}_k\mathbf{J}_{ik}^{\top}
    + (\sigma_{m,u}^2+\sigma_{m,c}^2)
      \mathbf{1}\mathbf{1}^{\top}.
  \label{eq:joint_covariance}
\end{equation}

Generalized least squares reduces the joint observation to a scalar
elevation and separates its variance into independent and potentially
correlated components:
\begin{equation}
  \begin{aligned}
    \mathbf{a}_{ik}
      &= \frac{\mathbf{R}_{ik}^{-1}\mathbf{1}}
      {\mathbf{1}^{\top}\mathbf{R}_{ik}^{-1}\mathbf{1}},
      \qquad
      \bar{z}_{ik}=\mathbf{a}_{ik}^{\top}\mathbf{z}_{ik}, \\
    r_{u,ik}
      &= \mathbf{a}_{ik}^{\top}\mathbf{D}_{ik}\mathbf{a}_{ik}
       + \sigma_{m,u}^2, \\
    r_{c,ik}
      &= \mathbf{a}_{ik}^{\top}\mathbf{J}_{ik}
         \boldsymbol{\Sigma}_k\mathbf{J}_{ik}^{\top}
         \mathbf{a}_{ik}
       + \sigma_{m,c}^2.
  \end{aligned}
  \label{eq:joint_measurement}
\end{equation}
The scalar observation $\bar{z}_{ik}$ has split variance
$r_{ik}=r_{u,ik}+r_{c,ik}$. The independent component $r_{u,ik}$
contains the projected stereo variance and $\sigma_{m,u}^2$. The
potentially correlated component $r_{c,ik}$ contains the propagated pose
variance and $\sigma_{m,c}^2$. Since
$\mathbf{a}_{ik}^{\top}\mathbf{1}=1$, each residual variance enters
the scalar observation once rather than decreasing with the number
of points.

For a newly observed cell, the estimate is initialized as
$\hat{h}_i=\bar{z}_{ik}$ and $p_{c,i}=r_{c,ik}$, while the total
variance satisfies $p_{u,i}+p_{c,i}=\max(r_{ik},\sigma_0^2)$.
Here, $\sigma_0^2$ sets the minimum initialization variance.
Each subsequent observation passes the consistency gate if
\begin{equation}
  \frac{(\bar{z}_{ik}-\hat{h}_i^{-})^2}
  {p_{u,i}^{-}+p_{c,i}^{-}+r_{ik}} < \gamma^2,
\end{equation}
where $\gamma=3$.

Although $\boldsymbol{\Sigma}_k$ contains the complete marginal pose
covariance at $t_k$, the external odometry interface does not provide
cross-covariances between the pose error at $t_k$ and errors at earlier
timestamps. Temporal correlations between $m_{c,ik}$ terms are likewise
unknown. Accepted observations are therefore fused using the scalar Split
Covariance Intersection (SCI)
update~\cite{julier2001simultaneous,li2013split}:
\begin{align}
  \tilde{p}_i(\omega)
    &= p_{u,i}^{-}+\frac{p_{c,i}^{-}}{\omega}, &
  \tilde{r}_{ik}(\omega)
    &= r_{u,ik}+\frac{r_{c,ik}}{1-\omega}, \\
  p_i^{+}(\omega)
    &= \left(\tilde{p}_i^{-1}
       +\tilde{r}_{ik}^{-1}\right)^{-1}, &
  \omega^{*}
    &= \underset{0<\omega<1}{\arg\min}\;p_i^{+}(\omega), \\
  \hat{h}_i^{+}
    &= p_i^{+}\left(
       \frac{\hat{h}_i^{-}}{\tilde{p}_i}
       +\frac{\bar{z}_{ik}}{\tilde{r}_{ik}}\right).
  \label{eq:sci_update}
\end{align}
All terms in the posterior mean are evaluated at $\omega^{*}$, and
the superscripts ${-}$ and ${+}$ denote prior and posterior estimates.
With $\lambda_i=p_i^{+}/\tilde{p}_i$ and
$\mu_i=p_i^{+}/\tilde{r}_{ik}$, the split is propagated as
$p_{u,i}^{+}=\lambda_i^2p_{u,i}^{-}+\mu_i^2r_{u,ik}$ and
$p_{c,i}^{+}=\max(p_i^{+}-p_{u,i}^{+},0)$.
Rejected observations leave the cell unchanged.
No process variance or observation aging is applied during the mapping
interval.

\subsection{Semantic Map Conversion and Application}
\label{sec:application}

This module integrates the probabilistic elevation map ${}^{G}\mathcal{M}_h$ with the structural map ${}^{G}\mathcal{M}_s$ for boundary and clearance estimation. Because the inputs are asynchronous, changed regions are converted periodically at 1 Hz.

\subsubsection{Semantic OctoMap Integration}

A colorized OctoMap ${}^{G}\mathcal{M}_o$ is formed by inserting ${}^{G}\mathcal{M}_s$ as \textit{structure} voxels and accepted cells from ${}^{G}\mathcal{M}_h$ as \textit{water}. Because LiDAR may occasionally capture weak and sparse returns from wave-induced ripples, ${}^{G}\mathcal{M}_h$ is used to classify the water surface during fusion. A structural point within $\delta_{ws}=0.1$ m of a valid water elevation is classified as a water return and excluded from the structure class. Water cells with uncertainty below $\sigma_{w,\max}$ are inserted, and previous water voxels are cleared at each update to preserve a 2.5D surface.

\subsubsection{2.5D Semantic Projection}

The semantic OctoMap is projected onto a 2.5D grid ${}^{G}\mathcal{G}_{\ell}$ with resolution $\Delta_c=0.4$ m. Each vertical column provides water elevation $z_w(i)$, local structural slope $s_i$ from a least-squares plane fit~\cite{fredriksson2024voxel}, and the lowest structural voxel $z_{s,\min}(i)$ above water. Their vertical gap is
$g_i = z_{s,\min}(i) - z_w(i)$. 

\subsubsection{Boundary Extraction and Clearance Estimation}

Each cell is labeled as \textit{water}, \textit{structure}, \textit{boundary}, or \textit{overhead}. Structural cells adjacent to water with $s_i>\tau_{\partial}=0.1$ define steep navigable boundaries. Cells with $g_i>1.5$ m are grouped into connected overhead regions $\mathcal{O}_r$~\cite{luo2025inland}. IHO chart specifications define bridge vertical clearance relative to the lowest part of the structure to obtain the minimum clearance for navigation~\cite{iho2026s4}. To reduce sensitivity to single reflected or misclassified cells, we report $C_{r,05}$, the 5th percentile of valid gaps in region $r$, as a lower-tail estimate. The labels and $C_{r,05}$ estimates are exported as semantic layers.

\section{EXPERIMENTS}
\subsection{Experimental Setup and Datasets}
\label{sec:exp_setup}

\begin{figure*}[!t]
  \centering
  \vspace*{8pt}
  \includegraphics[width=0.75\textwidth]{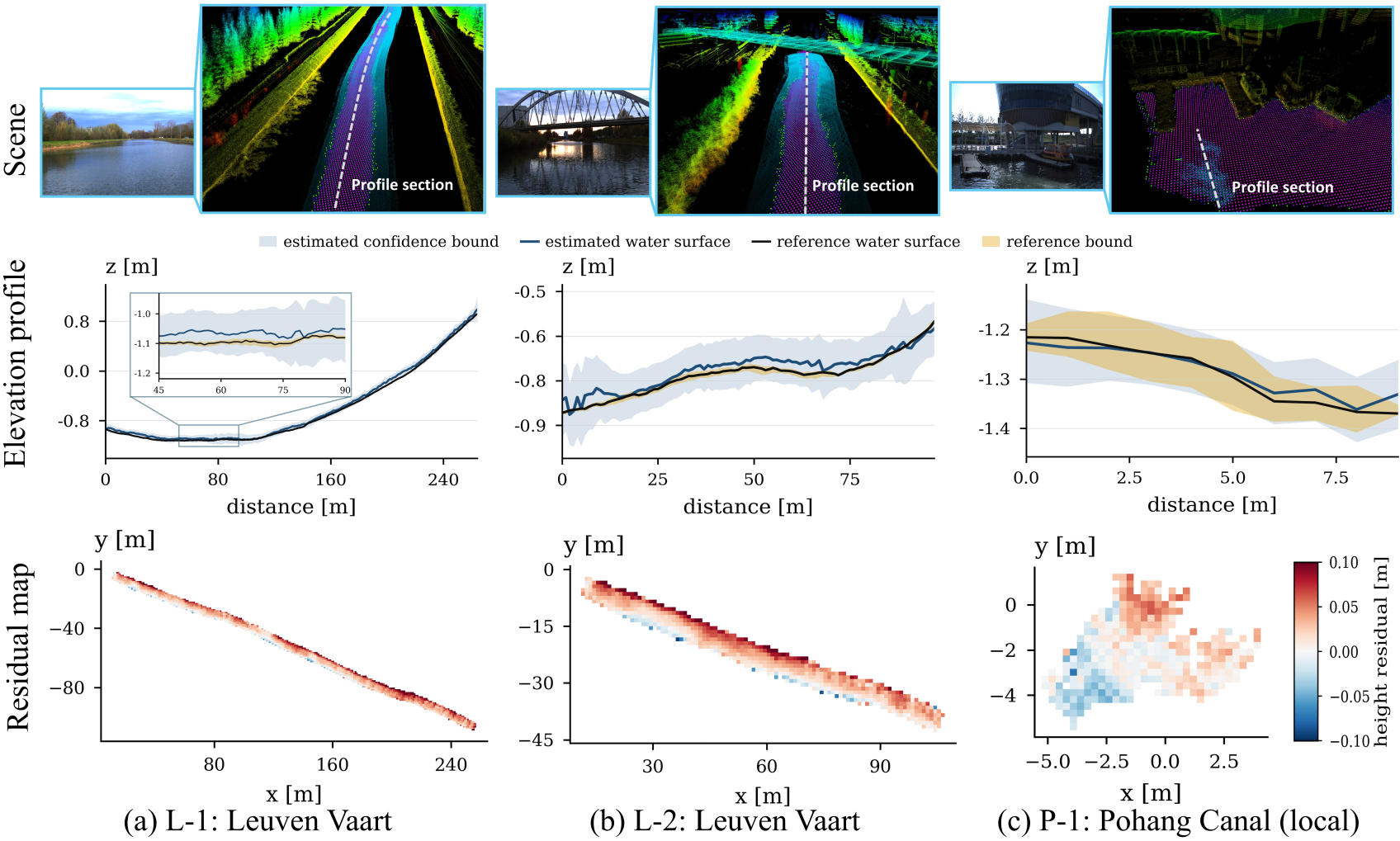}
  \caption{Water surface elevation results for L-1 and L-2 on Leuven Vaart and the local P-1 reference region on the Pohang Canal dataset. The top row combines representative images and registered 3D maps. Blue points denote the separated water surface, the remaining points are colored by vertical height, and the magenta surface denotes HydroMap. The middle row compares the longitudinal profiles with the estimated confidence bounds and reference bounds. The bottom row shows the spatial distribution of the elevation residuals.}
  \label{fig:water_profile_eval}
  \vspace*{-8pt}
\end{figure*}

We evaluate HydroMap using selected sequences from the public Pohang Canal dataset~\cite{chung2023pohang} and field experiments conducted on the Leuven Vaart. The Pohang dataset was collected in narrow waterways in Pohang, South Korea, and contains synchronized stereo images, LiDAR point clouds, global navigation satellite system (GNSS) measurements, and inertial measurement unit (IMU) data from narrow canal, port, and near-coastal scenes.

The Leuven field data were collected along the Leuven-Dijle Canal in Belgium using an instrumented catamaran equipped with a calibrated stereo camera pair, LiDAR, and GNSS/INS~\cite{zhang2026vessel}. During the Leuven experiments, the measured water turbidity near the acquisition site was 45--56 NTU. This turbidity level exceeds the threshold required for reliable LiDAR water surface detection~\cite{allis2011application}. Under these turbid conditions, visible LiDAR returns from the water surface were captured and used as reference observations for quantitative evaluation.

The selected sequences from both data sources are summarized in Table~\ref{tab:evaluation_sequences}. All experiments were processed on a laptop equipped with an Intel Core Ultra 9 275HX CPU, an NVIDIA GeForce RTX 5090 Laptop GPU, and 64 GB of RAM, running Ubuntu 22.04 via Windows Subsystem for Linux (WSL) and ROS 2. Together, the two data sources provide different vessel, sensor, and scene configurations for evaluating the framework.

\begin{table}[t]
  \centering
  \vspace*{8pt}
  \caption{Evaluation sequences.}
  \label{tab:evaluation_sequences}
  \setlength{\tabcolsep}{2.5pt}
  \renewcommand{\arraystretch}{1.08}
  \footnotesize
  \begin{tabularx}{\columnwidth}{@{}c>{\raggedright\arraybackslash}Xccc@{}}
    \toprule
    \textit{Seq.} & \textit{Dataset and scene} &
    \textit{Dist. [m]} & \textit{Water ref.} & \textit{Clearance} \\
    \midrule
    L-1 & Leuven Vaart, vegetated bank             & 285.8 & Yes     & -- \\
    L-2 & Leuven Vaart, bank with bridge           & 122.8 & Yes     & Measured \\
    P-1 & Pohang Canal, port area with bridges     & 233.5 & Partial & Estimated \\
    P-2 & Pohang Canal, canal with bridges         & 530.5 & No      & Estimated \\
    P-3 & Pohang Canal, canal with bridges         & 388.3 & No      & Estimated \\
    \bottomrule
  \end{tabularx}
  \par\vspace{3pt}
  \noindent\begin{minipage}[t]{\columnwidth}
    \scriptsize\raggedright
    \textit{Note:} Water ref. indicates the availability of LiDAR water surface returns used for quantitative evaluation. A dash indicates that the sequence contains no overhead structure requiring clearance evaluation. Measured indicates evaluation against an independent clearance measurement, whereas Estimated indicates that only the HydroMap estimate is reported.
  \end{minipage}
  \vspace*{-8pt}
\end{table}

\subsection{Water Segmentation Performance}
\label{sec:seg_performance}

Initial water masks were generated using SAM 2~\cite{ravi2025sam} with manually specified prompts. Each mask was then manually inspected and refined to ensure accurate separation between water and background, with particular attention to ambiguous water boundaries and regions affected by reflections, shoreline vegetation, or surrounding structures.

The test subsets were selected first to evaluate both segmentation and the complete mapping system; the remaining annotations were used for network development. The Pohang data comprise 1,698 images split into 697 training, 174 validation, and 827 test images. Of 252 Leuven images, 96 from distinct recording segments were reserved for testing and the remaining 156 for training and validation. Test images were excluded from optimization and checkpoint selection. In addition to the jointly trained model used for the mapping experiments, we trained models on Pohang and Leuven separately to evaluate cross-dataset transfer. Here, mIoU averages IoU over the water and background classes; $\mathit{IoU}_{w}$, $\mathit{P}_{w}$, $\mathit{R}_{w}$, and $\mathit{F1}_{w}$ denote water-class IoU, precision, recall, and F1 score, respectively.

\begin{table}[t]
  \centering
  \caption{Water segmentation performance across training and test datasets.}
  \label{tab:water_segmentation}
  \setlength{\tabcolsep}{1.5pt}
  \renewcommand{\arraystretch}{1.06}
  \scriptsize
  \begin{tabular*}{\columnwidth}{@{\extracolsep{\fill}}llccccc@{}}
    \toprule
    \textit{Training} & \textit{Test} & \textit{mIoU [\%]} & $\mathit{IoU}_{w}$ \textit{[\%]} &
    $\mathit{P}_{w}$ \textit{[\%]} & $\mathit{R}_{w}$ \textit{[\%]} & $\mathit{F1}_{w}$ \textit{[\%]} \\
    \midrule
    Pohang & Pohang & 98.54 & 97.90 & 98.59 & 99.29 & 98.94 \\
    Pohang & Leuven & 92.58 & 91.98 & 99.08 & 92.78 & 95.82 \\
    Leuven & Pohang & 82.66 & 76.94 & 77.04 & 99.84 & 86.97 \\
    Leuven & Leuven & 98.35 & 98.26 & 98.76 & 99.50 & 99.12 \\
    Pohang+Leuven & Pohang & 98.49 & 97.83 & 98.68 & 99.13 & 98.90 \\
    Pohang+Leuven & Leuven & 98.48 & 98.41 & 99.02 & 99.38 & 99.20 \\
    \bottomrule
  \end{tabular*}
\end{table}

Table~\ref{tab:water_segmentation} shows asymmetric cross-dataset transfer. The Pohang-trained model obtains $\mathit{IoU}_{w}=91.98\%$ on Leuven, whereas the Leuven-trained model obtains $76.94\%$ on Pohang. The latter combines $99.84\%$ recall with $77.04\%$ precision, indicating more false-positive water predictions. The jointly trained model gives $\mathit{IoU}_{w}=97.83\%$ on Pohang and $98.41\%$ on Leuven and is used for the mapping experiments.

\subsection{Water Surface Elevation Accuracy}
\label{sec:water_accuracy}

The water surface elevation was evaluated in verified regions containing LiDAR water returns. The returns were projected onto the HydroMap grid; the per-frame cell median and then the temporal median over at least five observations defined $h_i^{\mathrm{ref}}$. For cell error $e_i=\hat{h}_i-h_i^{\mathrm{ref}}$, we report bias (mean signed error), mean absolute error (MAE), root mean square error (RMSE), 95th percentile absolute error (P95), and empirical 95\% confidence interval coverage ($C_{95}$), defined as the fraction of reference elevations within $\hat{h}_i\pm1.96\sigma_{h,i}$. The estimated confidence bounds in Fig.~\ref{fig:water_profile_eval} visualize this uncertainty. The reference bounds span the 5th to 95th percentiles of the LiDAR elevations within each profile bin and do not enter the coverage calculation.

Table~\ref{tab:water_accuracy} shows centimeter-level differences from the LiDAR references in all three regions. The RMSE remains below $5$\,cm and P95 does not exceed $9.2$\,cm. In Fig.~\ref{fig:water_profile_eval}, the profiles follow the LiDAR references and the residual maps are dominated by magnitudes below $0.10$\,m. The estimated confidence bounds provide coverage close to the nominal level on L-1 and P-1 and slightly conservative coverage on L-2. The local P-1 reference was observed mainly during reverse motion, when the vessel wake roughened and aerated the water surface, exposing a wider range of incidence angles. Bubbles and foam can increase optical backscatter and reflectance, which may help explain the more frequently detectable LiDAR returns~\cite{zhang2004optical}. By contrast, persistent Leuven returns are consistent with near-surface scattering by suspended particles under the measured high turbidity~\cite{tamari2016stage}.

\begin{table}[t]
  \centering
  \caption{Water surface elevation accuracy.}
  \label{tab:water_accuracy}
  \setlength{\tabcolsep}{2.0pt}
  \renewcommand{\arraystretch}{1.08}
  \scriptsize
  \begin{tabular*}{\columnwidth}{@{\extracolsep{\fill}}ccccccc@{}}
    \toprule
    \textit{Seq.} &
    \textit{Cells} &
    \textit{Bias [cm]} &
    \textit{MAE [cm]} &
    \textit{RMSE [cm]} &
    \textit{P95 [cm]} &
    $C_{95}$ \textit{[\%]} \\
    \midrule
    L-1 & 16337 & 3.75 & 3.97 & 4.85 & 9.20 & 95.9 \\
    L-2 &  5071 & 1.88 & 2.94 & 3.95 & 8.10 & 98.1 \\
    P-1 &   327 & 0.53 & 2.04 & 2.57 & 5.25 & 92.7 \\
    \bottomrule
  \end{tabular*}
\end{table}

\subsection{Fusion Comparison}

SCI, the Kalman filter (KF), and covariance intersection (CI) receive the same joint cell observation from Eqs.~\eqref{eq:joint_covariance} and~\eqref{eq:joint_measurement}. KF assumes temporal independence and applies the conventional scalar Kalman update. For this comparison, CI~\cite{julier1997non} applies covariance intersection to the complete variance using the same symmetric weighting. The No Fusion baseline uses only the latest accepted cell observation. All other inputs and parameters are unchanged.

\begin{table}[t]
  \centering
  \vspace*{8pt}
  \caption{Fusion comparison. Values are ordered as L-1/L-2/P-1.}
  \label{tab:fusion_ablation}
  \setlength{\tabcolsep}{2.2pt}
  \renewcommand{\arraystretch}{1.06}
  \scriptsize
  \begin{tabular*}{\columnwidth}{@{\extracolsep{\fill}}lccc@{}}
    \toprule
    \textit{Method} & \textit{RMSE [cm]} & $C_{95}$ \textit{[\%]} & \textit{Med. $\sigma_h$ [cm]} \\
    \midrule
    SCI & 4.85/3.95/2.57 & 95.9/98.1/92.7 & 4.04/4.19/2.16 \\
    KF & 4.84/3.92/2.53 & 79.0/92.9/70.8 & 2.75/3.05/1.24 \\
    CI & 3.28/3.40/3.95 & 100/100/100 & 8.12/8.22/9.05 \\
    No Fusion & 3.39/4.08/5.70 & 99.9/99.8/100 & 8.09/8.10/9.06 \\
    \bottomrule
  \end{tabular*}
  \vspace*{-8pt}
\end{table}

SCI and KF yield nearly identical RMSEs, whereas CI and No Fusion differ more across regions; the clearest systematic separation is in the estimated uncertainty. Under the temporal independence assumption, KF reduces the median $\sigma_h$ to $1.24$--$3.05$\,cm, but its $C_{95}$ falls to $70.8$--$92.9\%$, indicating overly narrow intervals, particularly on P-1. SCI separates components assumed independent across observations from those with unavailable temporal cross-covariance. It gives a median $\sigma_h$ of $2.16$--$4.19$\,cm and coverage of $92.7$--$98.1\%$, closer to the nominal $95\%$ level. CI applies covariance intersection to the complete variance, whereas No Fusion does not combine observations; both produce intervals near $8$--$9$\,cm and almost $100\%$ coverage. SCI therefore retains centimeter-level RMSE across all three regions while reducing KF undercoverage and avoiding the wider intervals of CI and No Fusion.

\subsection{Semantic Mapping and Runtime}

\begin{figure}[!t]
  \centering
  \includegraphics[width=0.88\columnwidth]{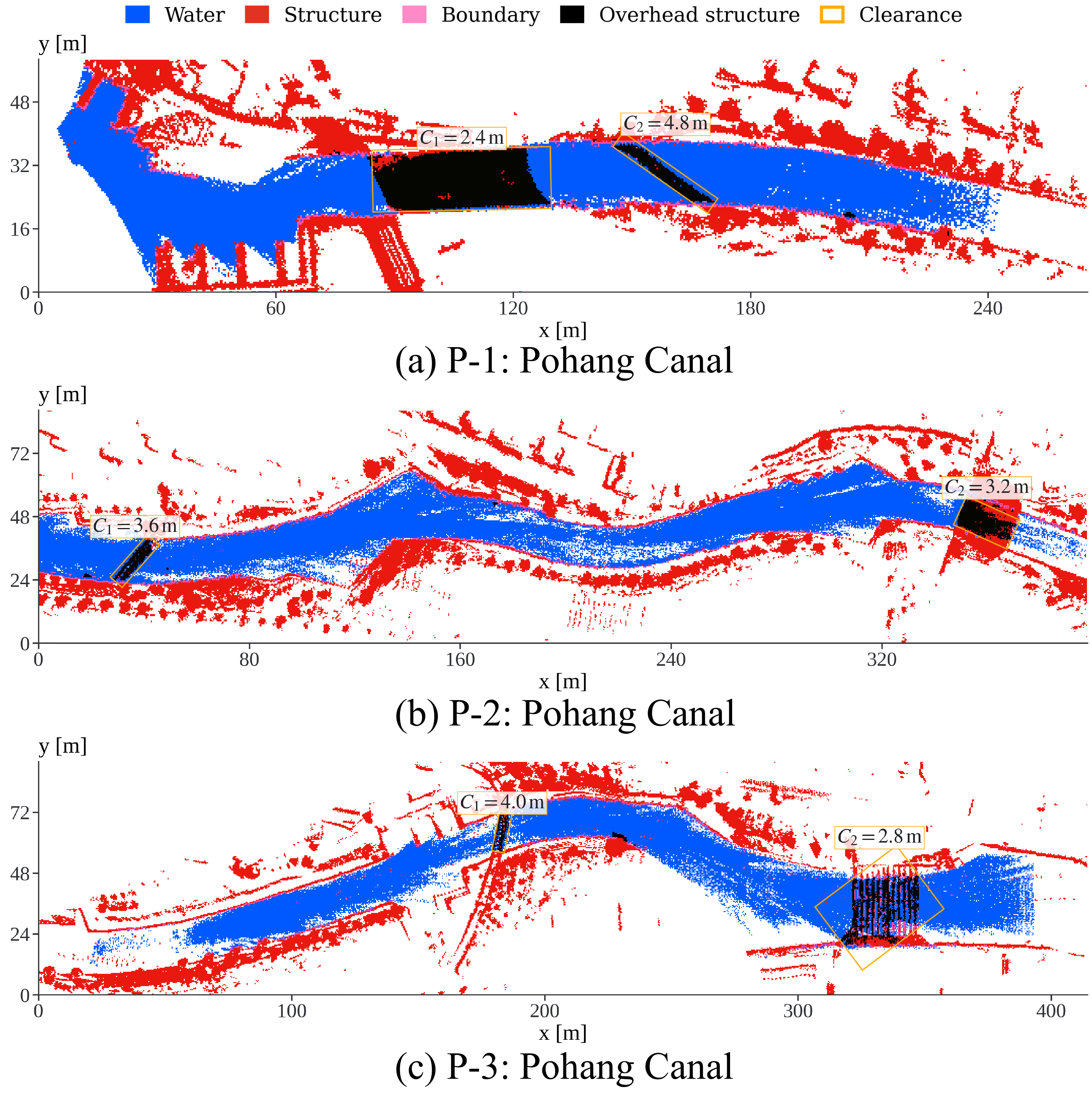}
  \caption{Semantic 2.5D maps and overhead clearance estimates for P-1, P-2, and P-3.}
  \label{fig:semantic_map_eval}
\end{figure}

\begin{figure}[!t]
  \centering
  \vspace*{8pt}
  \includegraphics[width=0.85\columnwidth]{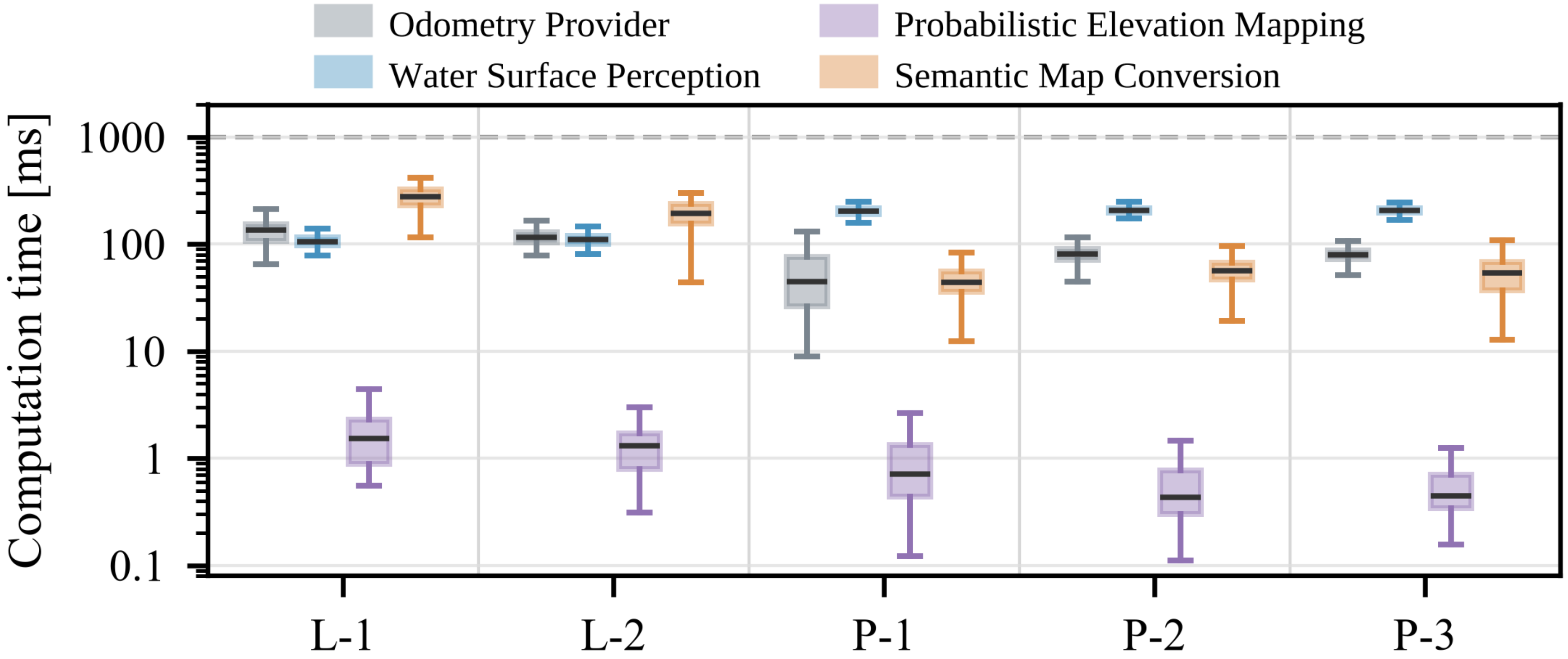}
  \caption{Runtime distributions of the four system modules across the evaluation sequences. The dashed line indicates the $1$\,s semantic conversion period.}
  \label{fig:runtime_eval}
  \vspace*{-8pt}
\end{figure}

In Fig.~\ref{fig:semantic_map_eval}, the semantic maps combine navigable water, boundaries, and overhead regions in a unified 2.5D representation. The $C_{r,05}$ clearance estimates for the six shown bridge regions in Pohang range from $2.4$ to $4.8$\,m. At L-2, HydroMap estimates $C_{r,05}=6.8$\,m for the independently measured bridge region, compared with the manual measurement of $6.85$\,m.

FAST-LIVO2 provides external odometry and is timed separately. HydroMap contains the remaining three modules, which are scheduled independently. Water Surface Perception runs in a separate process and uses the most recent stereo pair. Probabilistic Elevation Mapping fuses every accepted water observation and publishes the elevation map at $2$\,Hz, while Semantic Map Conversion updates the semantic map at $1$\,Hz. Fig.~\ref{fig:runtime_eval} therefore reports the computation time of one execution for each module.

In Fig.~\ref{fig:runtime_eval}, the median Probabilistic Elevation Mapping time remains below $1.6$\,ms. The external Odometry Provider and Water Surface Perception require $35$--$135$\,ms and $106$--$237$\,ms per execution, respectively. Semantic Map Conversion takes $39$--$56$\,ms in Pohang and $194$--$277$\,ms in Leuven. All sequences use the same OctoMap resolution. The Leuven experiments focus on water surface validation and use complete LiDAR structural clouds, whereas Pohang evaluates semantic navigable area mapping using points within the camera field of view. This broader input explains the longer Leuven conversion time. All measured executions finish within the $1$\,s semantic map update period.

\section{CONCLUSIONS, LIMITATIONS AND FUTURE WORK}

HydroMap reconstructs the water surface omitted by conventional LiDAR maps and integrates it with surrounding structures in a persistent semantic scene representation. It combines stereo perception with probabilistic temporal fusion while remaining decoupled from external odometry. Across three verified LiDAR reference regions, elevation RMSE remains below $5$\,cm and P95 does not exceed $9.2$\,cm. The semantic maps recover water boundaries and overhead regions, while the elevation and semantic maps are updated at $2$\,Hz and $1$\,Hz, respectively. Together, these outputs provide geometric inputs for navigable area delineation, overhead clearance assessment, and downstream planning.

Several limitations remain. Low texture, glare, mirror reflections, and low illumination can leave gaps in the stereo water layer or place reflected structure points in free space. The uncertainty model preserves covariance within each joint observation but omits dependence between neighboring cells and temporal pose cross-covariances; it also assumes a locally stationary surface. Moreover, HydroMap and its LiDAR reference share the FAST-LIVO2 trajectory, so the reported differences measure consistency in a shared map frame rather than absolute georeferenced accuracy. Clearance validation is limited to one manual measurement at L-2.

Future work will explore temporal or multiview completion, reflection rejection, and water surface models that account for waves, wakes, and changing water levels. We will also use independent surveyed references and repeated clearance measurements, integrate the semantic representation with planning and control, and investigate water surface constraints on vessel attitude and vertical drift.


\bibliographystyle{IEEEtran}
\bibliography{references}

\end{document}